# Industrial Steel Slag Flow Data Loading Method for Deep Learning Applications

**Authors**

Mert Sehri[1] (*), Ana Cardoso[2], Francisco de Assis Boldt[2], Patrick Dumond[1]

**Affiliations**

[1] Department of Mechanical Engineering, University of Ottawa, 161 Louis Pasteur, Ottawa, Ontario, Canada

[2] Department of Computer Science, Campus Serra do Instituto Federal do Espírito Santo (IFES), Vitória, Brazil

Corresponding author's email address and Twitter handle

**Email Address:** *msehr006@uottawa.ca, **Twitter handle:** @MMSehri
**Email Address:** franciscoa@ifes.edu.br
**Email Address:** pdumond@uottawa.ca
**Email Address:** abscardoso@gmail.com

**Abstract**

    Steel casting processes are vulnerable to financial losses due to slag flow contamination, making accurate slag flow condition detection essential. This study introduces a novel cross-domain diagnostic method using vibration data collected from an industrial steel foundry to identify various stages of slag flow. A hybrid deep learning model combining one-dimensional convolutional neural networks and long short-term memory layers is implemented, tested, and benchmarked against a standard one-dimensional convolutional neural network. The proposed method processes raw time-domain vibration signals from accelerometers and evaluates performance across 16 distinct domains using a realistic cross-domain dataset split. Results show that the hybrid convolutional neural network and long short-term memory architecture, when combined with root mean square preprocessing and a selective embedding data loading strategy,

achieves robust classification accuracy, outperforming traditional models and loading techniques. The highest test accuracy of 99.10 ± 0.30 demonstrates the method's capability for generalization and industrial relevance. This work presents a practical and scalable solution for real-time slag flow monitoring, contributing to improved reliability and operational efficiency in steel manufacturing.

## Nomenclature

AI     Artificial Intelligence
CNN   Convolutional Neural Network
DL     Deep Learning
LSTM Long Short-Term Memory
ML    Machine Learning
RNN   Recurrent Neural Network

## 1. Introduction

Steel manufacturing is a complex process vital to various sectors, like construction and transportation. The effective management of slag formation is a critical step in this process. Slag, a byproduct that forms during steel production, rises to the top of the molten steel, and needs to be separated. While slag is crucial in removing impurities and maintaining the purity of the final steel product, its management is complex and often inefficient. Improper management can lead to slag contamination, compromising steel quality and causing structural flaws. Therefore, accurate monitoring and controlling slag formation and separation is essential for producing high-quality steel while optimizing efficiency, minimizing waste, and meeting the demands of various industries [1].

Molten steel is poured from a ladle into a tundish through a refractory-lined ladle shroud. The ladle is designed to withstand hot temperatures and directs molten steel into the tundish, which serves as an intermediate container lined with refractory material. The flow between the ladle and tundish is controlled by lock and slide gates, with a second slide gate regulating flow into a water-cooled copper mold. A ladle shroud manipulator arm facilitates the removal of the shroud during ladle changes. During continuous casting, a slag layer forms atop the molten steel, removing impurities and insulating the steel from oxidation. As shown in Figure 1 [2].

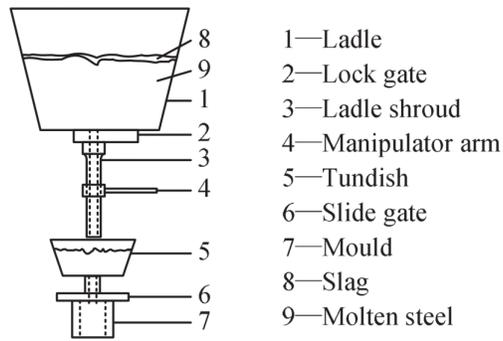

**Figure 1.** Part of the production flow of continuous casting machine [2]

Traditional methods for controlling slag in steel manufacturing are heavily reliant on operator experience and manual analysis, making them subjective, reactive, and prone to error [3]. Operators visually identify slag formation and manually control its removal, a process that struggles to adapt to the dynamic and complex conditions of steel casting. This often leads to inefficiencies, potential quality issues, and a reactive approach where problems are addressed only after they have impacted production. The complexity of slag formation further challenges traditional techniques, which often oversimplify the problem and hinder accurate predictions.

Recent advancements in artificial intelligence (AI) and machine learning (ML) are revolutionizing industrial processes, offering new ways to address challenges like those found in steel manufacturing. Vibration analysis using AI/ML has emerged as a powerful real-time monitoring tool. This innovative approach leverages vibration signals, captured by sensors during the steel casting process, to gain insights into the state of equipment and materials. AI/ML models can then be developed using this sensor data to predict and detect anomalies in the production line.

The main hypothesis of this work is that AI/ML models can accurately detect and classify various stages of slag flow by analyzing vibration data. These real-time insights give operators valuable information to make informed decisions. As a result, operators can ensure timely slag removal and prevent contamination, leading to improvements in efficiency and product quality.

For example, a study utilizing a combined CNN-LSTM architecture demonstrated the effectiveness of AI/ML models in classifying different slag flow conditions with high precision [4]. This approach is successful in detecting slag flow stages, including early, mid, and late phases, by analyzing vibration data collected from steel casting processes. The research emphasized the robustness of AI/ML models in managing the complex and dynamic environment of steel manufacturing, where real-time insights are critical. These advancements contribute to predictive maintenance strategies, ensuring timely interventions, optimizing slag removal, and enhancing overall operational efficiency and product quality in steel casting [5].

While existing research has demonstrated the potential of AI/ML applications in manufacturing to enhance efficiency and improve process control, particularly in areas like process condition inclusion prediction in continuous-casting processes, a gap remains in applying these techniques specifically to slag flow detection and classification in steel casting. This research aims to bridge this gap by developing a robust method that includes effective data division, data preprocessing, input data loading, and AI/ML model that can accurately predict and classify various stages of slag flow, contributing to a more efficient and sustainable steel manufacturing process.

The significance of this study lies in its potential to fill this critical gap in the literature on AI/ML applications in steel manufacturing by providing empirical evidence for the effectiveness of ML models in vibration-based slag detection. The study uses data from vibration sensors placed near the operator's hand on a mechanical arm used in steel casting. This data will be used to train a CNN-LSTM model to classify different stages of slag flow. Previous research, such as the work by Zhang et al [6], has already shown the effectiveness of ML algorithms in addressing complex industrial challenges. Building upon this foundation, this research aims to address the specific challenge of vibration-based slag detection, an area that has received limited attention in the existing literature. The results of this study could lead to more sustainable and efficient steel manufacturing processes, reducing waste, and improving overall product quality.

The remainder of the paper is divided into industrial set up, methodology, results, and conclusion.

## 2. Industrial Set Up

The steel slag flow dataset (SSFD) is collected at an industrial steel foundry to record flow changes in real time production. This allows to differentiate between molten metal and slag. Industrial set up of the SSFD is shown in Figure 2 where the data scientist placed the triaxial accelerometer sensor at the operator end of the casting process to capture vibration data without overheating the sensor. Flow conditions are recorded every 5 seconds at 6,400 Hz thus having 32,000 data samples for each set of data [7].

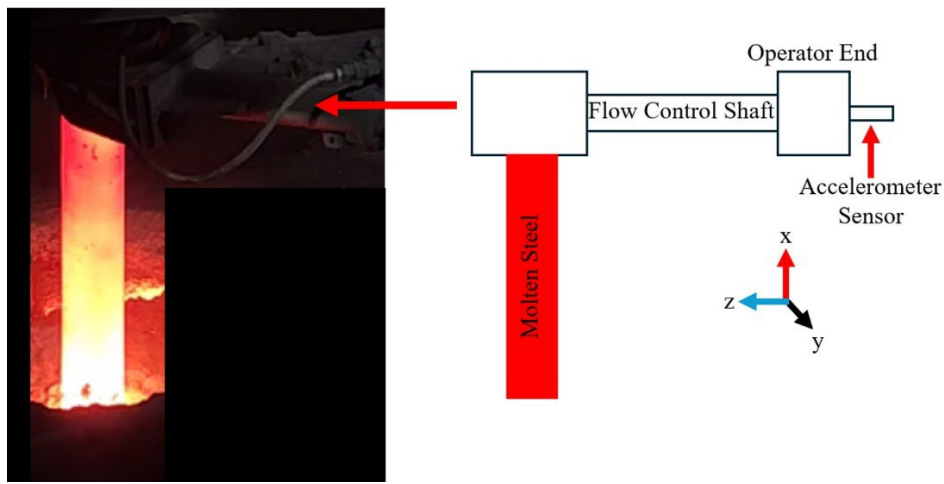

**Figure 2.** SSFD Industrial Set Up [7]

## 3. Methodology

### 3.1 Data Division

The proposed domain splitting of the dataset in Figure 3 makes it realistic for slag flow detection in real time monitoring conditions. Therefore, the SSFD offers practical applicability for steel slag flow detection, with insights gained from the study laying the groundwork for future industrial predictive maintenance practices. This research represents an initial step towards integrating ML research into industrial applications for slag flow condition monitoring.

Table 1 presents the domain splitting of the industrial SSFD. Each domain in the dataset consists of early slag flow, before slag flow, and during slag flow condition files. The data sample naming conventions consist of a letter representing the slag flow stage (E for early no slag, B for before slag, and S for during slag), followed by a number that indicates different conditions, ensuring diversity in testing. An example of a domain includes files like E-1, B-1, and S-1, representing various stages of slag flow in a single domain (domain name). The proposed cross-domain strategy is utilized by combining fifteen domains and testing on another domain to examine whether the trained model can transfer across domains for different slag flow conditions.

**Table 1.** Domain Splitting for the CWRU Dataset

| Domain Name | Early No Slag | Before Slag Flow | During Slag Flow |
|---|---|---|---|
| 1 | E-1 | B-1 | S-1 |
| 2 | E-2 | B-2 | S-2 |
| 3 | E-3 | B-3 | S-3 |
| 4 | E-4 | B-4 | S-4 |
| 5 | E-5 | B-5 | S-5 |
| 6 | E-6 | B-6 | S-6 |
| 7 | E-7 | B-7 | S-7 |
| 8 | E-8 | B-8 | S-8 |
| 9 | E-9 | B-9 | S-9 |
| 10 | E-10 | B-10 | S-10 |
| 11 | E-11 | B-11 | S-11 |
| 12 | E-12 | B-12 | S-12 |
| 13 | E-13 | B-13 | S-13 |
| 14 | E-14 | B-14 | S-14 |
| 15 | E-15 | B-15 | S-15 |
| 16 | E-16 | B-16 | S-16 |

SSFD proposed dataset splitting is provided in Figure 3.

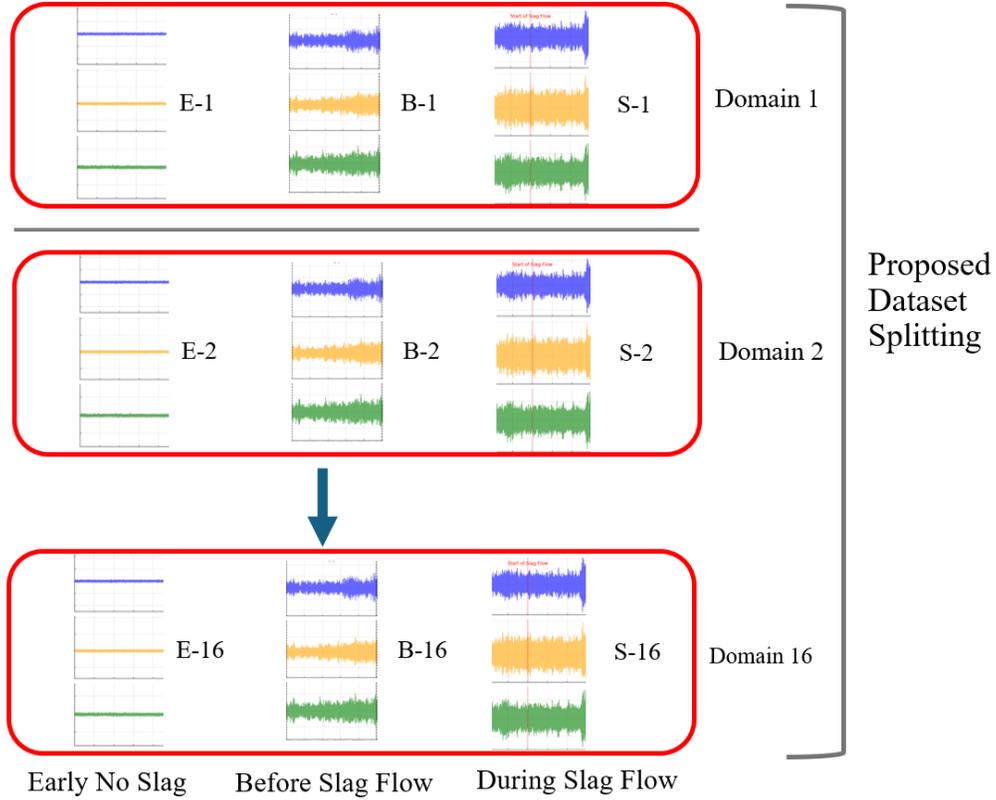

**Figure 3.** SSFD Proposed Dataset Splitting for Healthy and Developing Fault Bearings

### 3.2 Data Preprocessing

Preprocessing data is a crucial step in ML for flow classification, particularly in steel foundry applications like the SSFD [6], [7]. Despite the availability of various preprocessing techniques, a research gap remains in identifying which statistical methods can most effectively enhance the quality of sensor data for ML models, ensuring the reliability and accuracy of classification results. Many existing methods do not use straightforward statistical measures, such as standardization [8] and root mean square (RMS) [9], for normalizing datasets. This study includes basic statistical techniques to preprocess SSFD, making sure that the data fed into ML algorithm is normalized. Moreover, proper preprocessing reduces the computational time required for identifying slag flow.

The first preprocessing technique applied in this study is normalization through standardization (Z-score normalization), as outlined in equation (1). Where $x$ represents the individual data points being standardized, $\mu$ is the mean of the SSFD dataset file, and $\sigma$ is the standard deviation of the SSFD dataset file. Standardization adjusts the dataset to have a mean of 0 and a standard deviation of 1, creating a centralized baseline for comparison [8]. This is particularly important when assessing slag flow data, as it enables the detection of deviations that may indicate before or during slag flow conditions. By applying standardization, the SSFD can be compared across different conditions. This technique mitigates the impact of outliers and noise, which is essential for improving the performance of ML models. Standardization ensures that the features used for model training are on the same scale, enhancing the overall accuracy and efficiency of the classification process.

$$Z(x) = \frac{x - \mu}{\sigma} \quad (1)$$

The second preprocessing technique is the Root Mean Square (RMS), as described in equation (2), $n$ is the number of data points considered, the RMS used for normalization was not computed over the entire SSFD dataset. Instead, it was calculated only from the training dataset to ensure that no information from the test set was leaked during preprocessing. RMS is a statistical measure that accounts for both the magnitude and variability of data, making it particularly effective for detecting changes in vibration signals and identifying fault signatures. RMS is computed by taking the square root of the mean of the squared data points. This method amplifies larger deviations in the signal, making it easier to identify significant changes that could indicate more severe flow conditions. The application of RMS during preprocessing highlights the overall magnitude of a signal, a critical feature for flow classification. By reducing noise in the time-domain dataset, RMS helps to increase the robustness of ML models and mitigates the risk of overfitting.

$$RMS = \sqrt{\frac{\sum x^2}{n}} \quad (2)$$

There are several reasons why preprocessing raw SSFD using statistical methods like standardization and RMS is crucial. These techniques help minimize the impact of noise and outliers in the dataset, ensuring that ML models can better learn from the data. Effective preprocessing also helps to identify and correct inconsistencies in sensor readings, which can affect the performance of flow classification models. Furthermore, using simple statistical methods for preprocessing makes the process more accessible and easier to implement, providing a solid foundation that can support the integration of more complex techniques. Overall, a robust preprocessing method for SSFD is essential for ML-based fault classification in predictive maintenance to detect flow conditions.

### 3.3 Input Data Loading Strategies

Input data loading strategies are often overlooked by ML researchers as there are only a limited number of them. The traditionally utilized input data loading strategies by researchers are called traditional single-source data loading and traditional parallel loading. Traditional single-source data loading relies on input from a single sensor, where segments of data are loaded into a single-channel format. In contrast, traditional parallel data loading uses multiple sensors, loading each sensor's data separately in parallel. This setup enables the model to learn from diverse sources, improving generalizability and robustness. This paper compares the two traditional loading strategies and a third loading strategy proposed by Sehri et al. called selective embedding [10]. Selective embedding uses a single channel loading while utilizing multiple sensors loaded in an alternating fashion [10]. The three input data loading strategies is visualized in Figure 4.

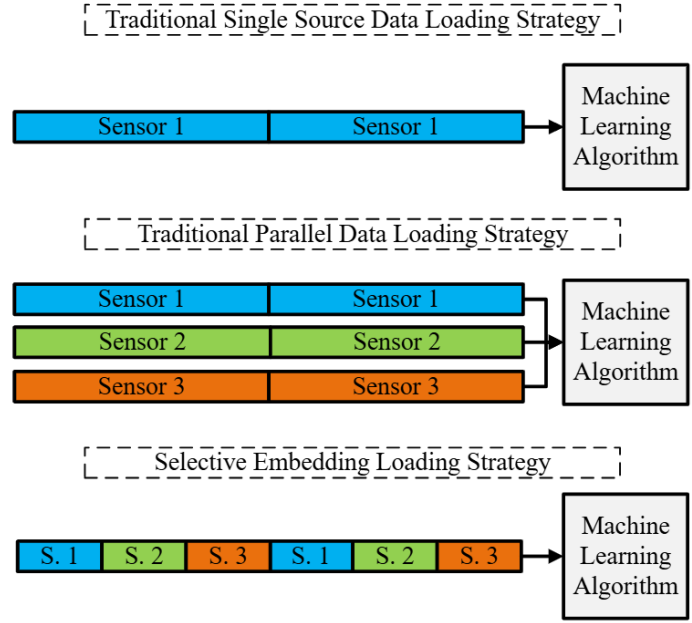

**Figure 4.** Input Data Loading Strategies [10]

### 3.4 AI/ML Model

Steel slag dataset represents a discrete random variable with a finite number of states, the data is categorized into three stages (early, before, and during slag) for detection. Due to this data division, neural networks are used for ML analysis [11]. The dataset includes labeled data that are related to different slag flow conditions, which are used to train the ML model.

To model the probability of each slag stage, a categorical distribution is used, where the random variable represents one of the three stages. This distribution is parameterized by a probability vector, $p \epsilon [0,1]^3$, which defines the likelihood of the system being in each slag stage given the observed vibration data. To estimate the probabilities [11], Equation 3 and 4 are used.

$$P(x = k) = p_k, k\epsilon\{early, before, during\} \qquad (3)$$

Where $p_k$ represents the probability of being in stage $k$.

$$\sum_k P(x = k) = 1 \qquad (4)$$

Convolutional neural networks (CNNs) are a type of deep learning (DL) model specifically designed for processing structured data, such as images [12]. CNNs consist of multiple layers, including convolutional layers, pooling layers, and fully connected layers [13]. Convolutional layers filter the input data, identifying patterns such as edges and textures in spatial features [14]. Pooling layers reduce the dimensionality of the data, making the network more efficient for processing smaller inputs, thus enabling CNNs to effectively learn features from raw data [15].

For this research Table 2 shows the CNN network architecture used for slag flow conditions testing.

**Table 2.** Proposed CNN Architecture

| Layers | Structures |
|---|---|
| 1 | Conv1d(1, 16, kernel_size=15), BatchNorm1d(16), ReLU |
| 2 | Conv1d(16, 32, kernel_size=3), BatchNorm1d(32), ReLU, MaxPool1d(kernel_size=2, stride=2) |
| 3 | Conv1d(32, 64, kernel_size=3), BatchNorm1d(64), ReLU |
| 4 | Conv1d(64, 128, kernel_size=3), BatchNorm1d(128), ReLU, AdaptiveMaxPool1d(4) |
| 5 | Linear(128 * 4, 256), ReLU, Dropout |
| 6 | Linear(256, 256), ReLU, Dropout |
| 7 | Linear(256, 10) |

On the other hand, recurrent neural networks (RNNs) are designed to handle sequential data like time-series information. RNNs have hidden states that store information from previous time steps, allowing them to capture temporal dependencies in the data [15]. This makes RNNs ideal when the order of data points is critical, such as in time-series analysis. However, training RNNs can be challenging due to vanishing or exploding gradients, which can hinder learning over long sequences [15]. To overcome this, more advanced variants like long short-term memory (LSTM) networks have been developed [16], enabling RNNs to learn long-term dependencies in the data more effectively.

The proposed DL method consists of a combination of network architecture, dataset division, and data preprocessing. Starting with a combined 1D CNN-LSTM architecture which is proposed (outlined in Table 3) to classify three classes of slag flow conditions: early no slag, before slag, and during slag. This model, inspired by prior condition monitoring algorithms, processes time-domain data from steel slag flow sensors. The algorithm includes a series of Conv1D layers paired with max-pooling layers to extract hierarchical features from the input signals. The convolutional layers use distinct filter sizes and depths, with ReLU activation functions, to detect patterns in the input data. The output is then flattened and passed through two fully connected layers, one of which contains 9 kernels with ReLU activation, producing a probability distribution over the sixteen slag flow conditions. The model aims to learn patterns from the input data through convolutional operations and is trained using categorical cross-entropy loss to validate various stages of slag flow across domains.

Vibration data is utilized in this study. Due to the robustness of the CNN and LSTM layers within this multi-modal deep learning model, the steel slag flow stages early no slag, before slag, and during slag are effectively extracted using cross-domain evaluation. The CNN-LSTM layers are specifically designed for processing vibration data, based on small dataset size to reduce computation time while maintaining high performance. For this experiment Table 3 shows the proposed CNN-LSTM architecture.

**Table 3.** Proposed CNN-LSTM Architecture

| Layers | Structures |
|---|---|
| 1 | Conv1d(1, 32, kernel_size=5, padding=3), BatchNorm1d(32), ReLU |
| 2 | Conv1d(32, 64, kernel_size=3, padding=2), BatchNorm1d(64), ReLU, MaxPool1d(kernel_size=2, stride=2) |
| 3 | Conv1d(64, 128, kernel_size=3, padding=1), BatchNorm1d(128), ReLU |
| 4 | Conv1d(128, 256, kernel_size=3, padding=1), BatchNorm1d(256), ReLU, AdaptiveMaxPool1d(1) |
| 5 | LSTM(256, 100, num_layers=3, batch_first=True, dropout=0.5, bidirectional=True) |
| 6 | Linear(200, 512), BatchNorm1d(512), ReLU, Dropout(0.5) |
| 7 | Linear(512, 256), BatchNorm1d(256), ReLU, Dropout(0.5) |
| 8 | Linear(256, 3) |

The framework proposed by Zhao et al. has been adapted to develop a ML model that optimizes time-domain data, allowing to compare cross-domain accuracies in slag flow conditions for steel foundry applications [17]. This approach strengthens flow detection and classification, improving predictive maintenance efficiency in steel casting operations.

For this study, the SSFD is used and organized for analysis [7]. This dataset's naturally occurring material flow cases make it valuable for real-world industrial conditions and enhancing its robustness. The statistical methods selected for preprocessing enrich the features of sensor data, thus improving the detection of slag flow states. Diverse conditions, and unknown mixture of steel and other materials are necessary in the dataset for statistical preprocessing techniques like standardization and RMS to be effective.

## 4. Results

The results involve testing various domains using cross-domain analysis with the SSFD. Each experiment is run 10 times, with the average accuracy of the best epoch taken for each run. A total of 100 epochs are evaluated, with a learning rate set to 0.001 for all cases. The experiments are performed using Python 3.10.9 and PyTorch 1.12.0, installed via conda. All ML models are trained on an NVIDIA GeForce RTX 3070 GPU, utilizing CUDA version 11.3 and CUDNN 8.1. The machine running the experiments is powered by a Windows 11 operating system, with an Intel Core i9-12900H CPU, 1TB SSD, and 32GB of RAM. The key Python libraries used in the experiments include numpy 1.23.4, matplotlib 3.6.3, pandas 1.5.0, scikit-learn 1.1.2, torch 2.3.1+cu118, tqdm 4.64.1, torchvision 0.18.1+cu118, torchaudio 2.3.1+cu118, and datasets 2.19.1.

### 4.1 Accelerometer Data Preliminary Results for Before Slag and During Slag Flow

The domain loading for training, validation and testing for SSFD dataset is shown in Table 4. In the preliminary experiments, to determine the best preprocessing and hyperparameters, the domains 1, 2, 3, 4, 5, 6, 7, 8, 9, 10, 11, 12, 13, 14, and 15 are used for training and validation and domain 16 is used for testing.

**Table 4.** SSFD Dataset Domain Data Train, Validation and Test Loading

| Domain Train Validation | Domain Tested |
|---|---|
| Every domain except 16 | 16 |
| Every domain except 15 | 15 |
| Every domain except 14 | 14 |
| Every domain except 13 | 13 |
| Every domain except 12 | 12 |
| Every domain except 11 | 11 |
| Every domain except 10 | 10 |
| Every domain except 9 | 9 |
| Every domain except 8 | 8 |
| Every domain except 7 | 7 |
| Every domain except 6 | 6 |
| Every domain except 5 | 5 |
| Every domain except 4 | 4 |
| Every domain except 3 | 3 |
| Every domain except 2 | 2 |
| Every domain except 1 | 1 |

Table 5 showcases the performance of different ML architectures and hyperparameters on the SSFD using accelerometer data. The proposed 1D CNN-LSTM model demonstrates strong stability, with test accuracies exceeding 65% in most cases, particularly when using RMS preprocessing. The best performance is observed with a batch size of 64 and an input signal length of 512, achieving a test accuracy of 82.76% ± 2.91. These findings suggest that the model configuration with RMS preprocessing, a batch size of 64, and an input signal length of 512 offers the best balance between accuracy and stability, making it the ideal configuration for further accelerometer data testing in the steel slag flow context.

**Table 5.** Accelerometer Data Results for Different ML Models and Hyperparameters for SSFD using Before Slag and During Slag Flow (Y- axis)

| Model Type | Preprocessing | Domain Train Validation | Domain Tested | Train Accuracy (%) | Validation Accuracy (%) | Test Accuracy (%) | Batch Size | Input Signal |
|---|---|---|---|---|---|---|---|---|
| 1D CNN | Standardization | 1,2,3,4,5,6,7,8,9,10,11,12,13,14,15 | 16 | 99.40 | 99.67 | 76.69 ± 2.96 | 64 | 512 |
| | | | | 99.35 | 99.59 | 68.06 ± 2.26 | 64 | 1024 |
| | | | | 99.87 | 99.13 | 71.00 ± 4.48 | 64 | 2048 |
| | | | | 99.93 | 99.41 | 77.58 ± 1.76 | 128 | 512 |
| | | | | 99.88 | 99.64 | 65.16 ± 6.17 | 128 | 1024 |
| | | | | 99.37 | 99.65 | 66.33 ± 5.47 | 128 | 2048 |
| | RMS | 1,2,3,4,5,6,7,8,9,10,11,12,13,14,15 | 16 | 99.71 | 99.15 | 77.66 ± 1.49 | 64 | 512 |
| | | | | 99.33 | 99.38 | 69.03 ± 3.80 | 64 | 1024 |
| | | | | 99.70 | 99.92 | 66.00 ± 4.42 | 64 | 2048 |
| | | | | 99.66 | 99.13 | 76.37 ± 2.72 | 128 | 512 |
| | | | | 99.89 | 99.60 | 66.77 ± 4.34 | 128 | 1024 |
| | | | | 99.78 | 99.16 | 66.00 ± 4.16 | 128 | 2048 |
| 1D CNN-LSTM | Standardization | 1,2,3,4,5,6,7,8,9,10,11,12,13,14,15 | 16 | 99.93 | 99.55 | 82.06 ± 2.98 | 64 | 512 |
| | | | | 99.36 | 99.76 | 65.64 ± 3.54 | 64 | 1024 |
| | | | | 99.69 | 99.99 | 56.33 ± 5.47 | 64 | 2048 |
| | | | | 99.36 | 99.62 | 81.77 ± 3.87 | 128 | 512 |
| | | | | 99.37 | 99.62 | 65.65 ± 2.98 | 128 | 1024 |
| | | | | 99.55 | 99.38 | 57.41 ± 4.09 | 128 | 2048 |
| | RMS | 1,2,3,4,5,6,7,8,9,10,11,12,13,14,15 | 16 | 99.69 | 99.34 | 82.76 ± 2.91 | 64 | 512 |
| | | | | 99.11 | 99.68 | 63.55 ± 5.82 | 64 | 1024 |
| | | | | 99.63 | 99.82 | 55.67 ± 3.00 | 64 | 2048 |
| | | | | 99.33 | 99.61 | 81.45 ± 3.75 | 128 | 512 |
| | | | | 99.79 | 99.82 | 65.32 ± 2.31 | 128 | 1024 |
| | | | | 99.93 | 99.67 | 57.33 ± 5.54 | 128 | 2048 |

## 4.2 Ablation Study

Ablation experiments are conducted for the proposed hybrid deep learning architecture. Table 6 includes the summary of experiments, where model components and configurations are assessed for the SSFD dataset. Ablation study allowed for the assessment of removing features to see which feature had the most effect on testing accuracy. Figure 5 illustrates the domain test accuracy results for methods A1 through A8 and the proposed methods M9 and M10, across 16 test domains. The proposed method, M9 and M10, achieve the highest average accuracies of 99.10 ± 0.30 and 93.56 ± 2.23, outperforming the other methods. For example, methods A2 and A6 show lower accuracies of 63.13 ± 2.00 and 61.76 ± 1.67, respectively, indicating that excluding features negatively impacts the model's performance.

The box plot highlights the accuracy distribution and standard deviations for each method. The proposed method M9 demonstrates a high median accuracy with a narrow variance, showcasing its robustness and effectiveness. Conversely, methods such as A2, A3, and A6 display wider variance and lower median values, underscoring their reduced consistency and generalization capability. These results emphasize that excluding certain features degrades performance, while including all features in the proposed methods enhances accuracy and generalization.

**Table 6.** Ablation Experiments for Proposed Method Early, Before, and During Slag

| Method | CNN Layers | LSTM Layers | RMS Preprocessing | Accelerometer Data X-Axis | Accelerometer Data Y-axis | Accelerometer Data Z-axis | Note |
|---|---|---|---|---|---|---|---|
| A1 | ✓ |  | ✓ | ✓ |  |  | CNN-LSTM, with single data, batch size 64, input signal 512. Training included before, and during slag. Single channel loading. |
| A2 | ✓ |  | ✓ |  | ✓ |  | CNN-LSTM, with single data, batch size 64, input signal 512. Training included before, and during slag. Single channel loading. |
| A3 | ✓ |  | ✓ |  |  | ✓ | CNN-LSTM, with single data, batch size 64, input signal 512. Training included before, and during slag. Single channel loading. |
| A4 | ✓ | ✓ | ✓ | ✓ |  |  | LSTM only, with single data. Training included before, and during slag. Single channel loading. Single channel loading. |
| A5 | ✓ | ✓ | ✓ |  | ✓ |  | CNN-LSTM, with preprocessing. Training included before, and during slag. Single channel loading. Single channel loading. |
| A6 | ✓ | ✓ | ✓ |  |  | ✓ | CNN-LSTM, with preprocessing. Training included before, and during slag. Single channel loading. Single channel loading. |
| A7 | ✓ | ✓ |  |  | ✓ |  | CNN-LSTM, without preprocessing. Training included early, before, and during slag. batch size 64, input signal 512. Single channel loading. |
| A8 | ✓ | ✓ | ✓ |  | ✓ |  | CNN-LSTM, with single data, batch size 64, input signal 512. Training included early, before, and during slag. Single channel loading. |
| M9 | ✓ | ✓ | ✓ | ✓ | ✓ | ✓ | Proposed method (baseline model), batch size 64, input signal 512. Training included before, and during slag. Single channel selective embedding [10]. |
| M10 | ✓ | ✓ | ✓ | ✓ | ✓ | ✓ | Proposed method (baseline model), batch size 64, input signal 512. Training included before, and during slag. Multi channel parallel loading. |

For the combined x, y, and z-axis SSFD results, the model performance improves when data from all three input axes are used as a multi channel input data parallel loading and fed into the DL architecture (M10).

Further testing is conducted, A4 (x-axis), A5 (y-axis), and A6 (z-axis) for different domains using the best-selected parameters from previous evaluations. For the x-axis SSFD, domains 1, 8, 9, and 12 show the lowest test accuracy results, with domain 12 achieving a particularly low result of $51.61 \pm 2.87$ and domain 9 showing $59.92 \pm 2.96$. Similarly, domains 1 and 8 exhibit lower test accuracy at $57.18 \pm 4.12$ and $58.73 \pm 1.96$, respectively, indicating instability in the model across these specific domains. The primary challenge remains improving the model by enhancing feature extraction techniques to achieve higher accuracies with lower standard deviations, thereby improving cross-domain slag flow detection.

For the y-axis SSFD, similar issues are present in particular domains. Notably, domain 12 stands out with a low-test accuracy of $53.82 \pm 3.41$. Domains 7 and 9 also show subpar results, with test accuracies of $58.17 \pm 2.88$ and $61.23 \pm 2.45$, respectively. These results suggest that the model struggles to generalize across different domains, particularly those associated with varying conditions during slag flow, indicating that the y-axis data might require more robust preprocessing or different feature extraction methods to improve accuracy.

For the z-axis SSFD, the model performance also shows variability across domains. Domain 12, for instance, presents a low-test accuracy of $51.73 \pm 1.54$, while domain 8 follows with an accuracy of $53.11 \pm 2.00$. Domain 4 also yields a low result at $59.32 \pm 1.78$. These results highlight the ongoing challenges with achieving consistent performance across different domains and underline the need for further improvement in feature extraction techniques and model training to enhance the model's robustness for real-world slag flow detection applications.

From the Figure 5 results for A4, A5, and A6, it is evident that the y-axis yielded the best performance among the individual axes. Therefore, only the y-axis and the combined x, y, and z axes are tested for more in-depth analysis of three conditions to determine their effectiveness in slag flow detection. Additionally, the results for A8 when using y-axis vibration data with RMS preprocessing across the early no slag, before slag flow, and during slag flow stages using the 1D CNN-LSTM model. For y-axis results the overall test accuracy across these domains averages 50.68 ± 2.93, reflecting a challenge in distinguishing between the early no slag and before slag flow stages due to the similarity in the data. This overlap in conditions contributes to reduced accuracy and increased variability. Therefore, researchers are encouraged to either focus on early no slag or before slag flow data when combining with during slag flow data to improve detection accuracy in steel casting applications. This adjustment would lead to better classification and detection in the context of steel slag flow processes.

The multi channel parallel input data loading approach for before and during slag yields an average test accuracy of 93.56 ± 2.23 (M10), indicating that integrating data in multiple channels from all three axes helps the model generalize features and improves its ability to generalize across different domains while a single channel selective embedding strategy [10] reaches average accuracy of 99.10 ± 0.30 proving that a single channel data loading can be more efficient while able to generalize better with preprocessing techniques combined. This result highlights the potential of utilizing a multi-axis input strategy to enhance the robustness and reliability of slag flow detection, emphasizing the importance of capturing comprehensive spatial information for effective slag flow detection. However, a domain still show variability in accuracy, indicating that further optimization in the preprocessing and feature extraction methods is needed for consistent detection performance.

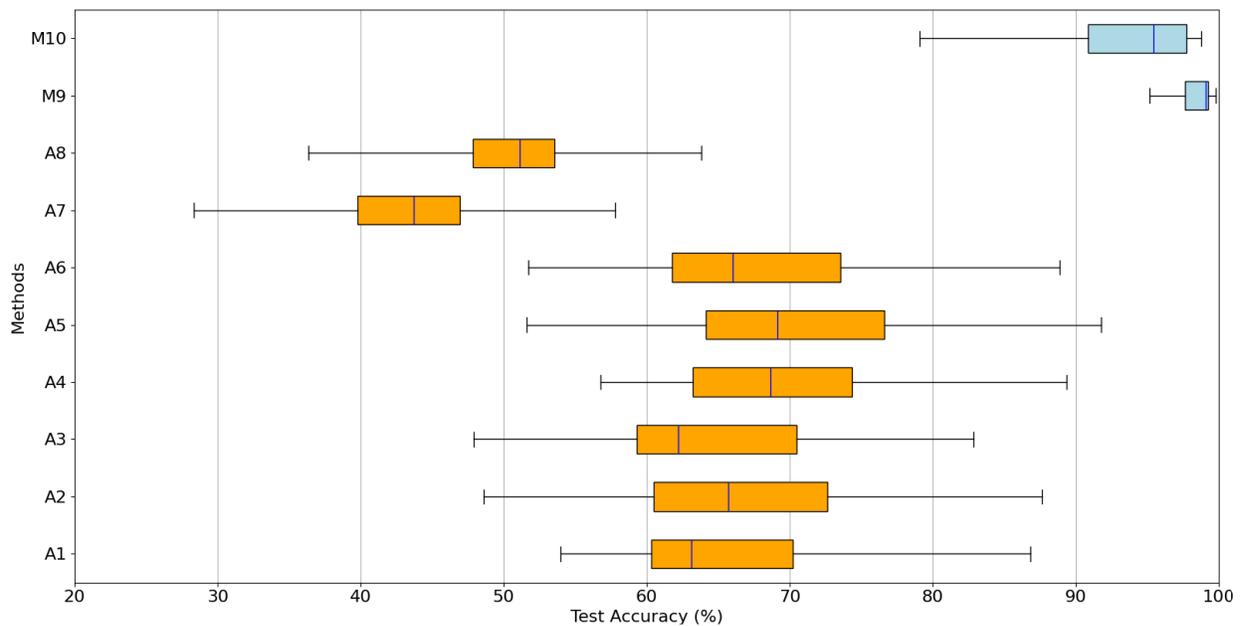

**Figure 5**. Box Plot of Ablation Study Results

Figure 6 shows the cross-domain results using confusion matrices for the tested domains 12, 13, 14, and 15. These confusion matrices highlight the model's ability to correctly classify slag flow stages across different domains for M9. Notably, the model performed well in distinguishing 'No-Slag' and 'Slag' states, with minimal misclassification errors. However, there is still mislabelling between 'Slag' and the other two categories, particularly for domain 12, where a higher number of samples from the 'Slag' stage are incorrectly classified as 'No Slag.' This emphasizes the need for further improvements in capturing and distinguishing temporal characteristics of the various stages, by enhancing the sequence modeling capacity of the LSTM component in the architecture.

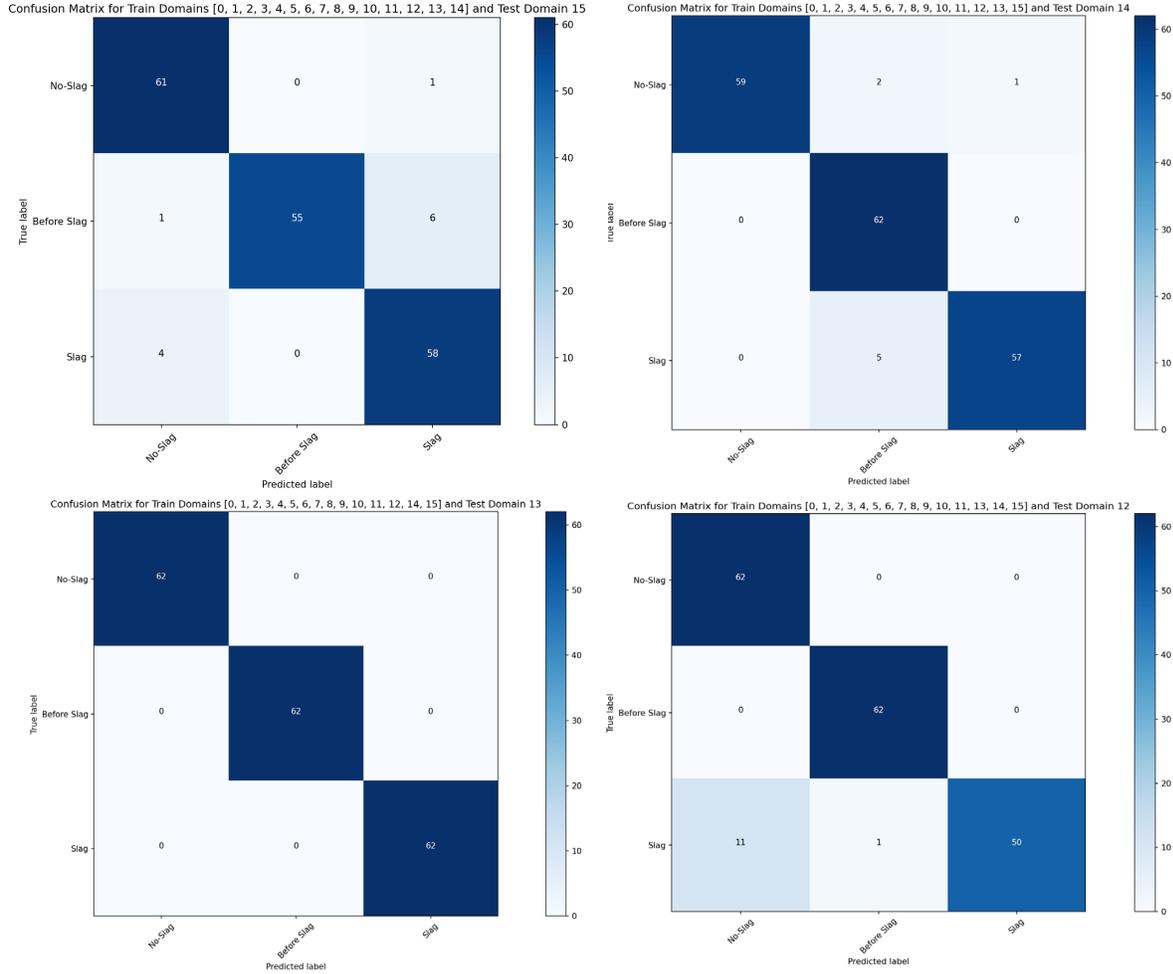

**Figure 6.** Cross Domain Results for Tested Domains 12,13,14,15

## 5. Conclusion

A novel method (M9) to data division for detecting steel slag flow conditions through the application of CNN-LSTM models on the SSFD, which captures various stages of slag flow under diverse operating conditions is proposed. By utilizing and preprocessing raw time-domain data from accelerometer signals, the proposed methodology demonstrates improvements in slag flow detection accuracy and model stability compared to traditional approaches like standalone 1D-CNNs.

The results highlight the practical value of the SSFD as a foundational step towards accurate slag flow detection in cross-domain applications. The proposed methods achieved a test accuracy of 99.10 ± 0.30 (M9) and 93.09 ± 2.50 (M10), which is a strong result for industrial applications, especially when considering the complex and dynamic nature of steel slag flow during production. This performance, tested across sixteen distinct cases, suggests the reliability and robustness of the flow detection process in real-world steel foundry environments.

Future work will focus on optimizing the model further to achieve even higher accuracy and stability for effective slag flow condition monitoring. The continued development of these techniques shows potential for improving operational efficiency in industrial steel casting applications.